\documentclass[conference]{IEEEtran}
\IEEEoverridecommandlockouts
\usepackage{cite}
\usepackage{amsmath,amssymb,amsfonts}
\usepackage{algorithmic}
\usepackage{graphicx}
\usepackage{textcomp}
\usepackage{xcolor}
\usepackage{multirow}
\usepackage{hyperref}
\def\BibTeX{{\rm B\kern-.05em{\sc i\kern-.025em b}\kern-.08em
    T\kern-.1667em\lower.7ex\hbox{E}\kern-.125emX}}
\begin{document}

\title{q-RBFNN: A Quantum Calculus-based RBF Neural Network}

\author{Syed Saiq Hussain, Muhammad Usman, Taha Hasan Masood Siddique,\\ Imran Naseem, Roberto Togneri and Mohammed Bennamoun   
 \thanks{Syed Saiq Hussain is with the College of Engineering, Karachi Institute of Economics and Technology, Karachi, Pakistan, e-mail: saiqhussain@gmail.com, Muhammad Usman is with the Department of Computer Engineering, Chosun University, Gwangju 61452, South Korea, e-mail: usman@chosun.kr, Taha Hasan Masood Siddique is with the Department of Electronic Engineering, NEDUET, e-mail: 1tahahassan1@gmail.com, Imran Naseem is with the College of Engineering, Karachi Institute of Economics and Technology, Karachi, Pakistan and School of Electrical, Electronic and Computer Engineering, The University of Western Australia, e-mail: imrannaseem@pafkiet.edu.pk, imran.naseem@ee.uwa.edu.au, Roberto Togneri is with the School of Electrical, Electronic and Computer Engineering, The University of Western Australia, e-mail: roberto.togneri@uwa.edu.au,  Mohammed Bennamoun is with the School of Computer Science and Software Engineering, The University of Western Australia, e-mail: mohammed.bennamoun@uwa.edu.au.}
  \thanks{*This is a preprint version.}}

\maketitle

\begin{abstract}
In this research a novel stochastic gradient descent based learning approach for the radial basis function neural networks (RBFNN) is proposed.  The proposed method is based on the $q$-gradient which is also known as Jackson derivative. In contrast to the conventional gradient, which finds the tangent, the $q$-gradient finds the secant of the function and takes larger steps towards the optimal solution. The proposed $q$-RBFNN is analyzed for its convergence performance in the context of least square algorithm. In particular, a closed form expression of the Wiener solution is obtained, and stability bounds of the learning rate (step-size) is derived.  The analytical results are validated through computer simulation. Additionally, we propose an adaptive technique for the time-varying $q$-parameter to improve convergence speed with no trade-offs in the steady state performance. The proposed time variant $q$-RBFNN has shown superior performance on the following tasks: (1) non-linear system identification problem, (2) hammerstein model, (3) multiple input multiple output (MIMO) system, and (4) chaotic time series prediction. The MATLAB implementation of the proposed method is available at the author’s GitHub page (https://github.com/musman88/q-RBFNN).
\end{abstract}


\section{Introduction} \label{intro}
The advent of advanced machine learning algorithms has demonstrated tremendous performances for the challenging problems \cite{watt2020machine}. The intricate machine learning methods including deep learning, support vector machines and random forest are now greatly employed in a variety of scientific applications \cite{usman2019afp,usman2020afp}. This development, however, is achieved with large amounts of training data and at the cost of computation overhead that is required for the training of these algorithms \cite{jan2019deep}. Moreover, the architecture of these algorithms is equally complex and therefore, the cost of implementation to user is substantially increased. On the other hand, the classical neural networks have simpler architecture and significantly lower implementation cost with agreeable performance. Specially, when the task is simple and available training data is scarce \cite{min2019shallow}.

For instance, radial basis function neural network (RBFNN) with its virtue of simple single layered architecture and its universal approximation capabilities has been adopted in a broad range of applications \cite{rahmati2019application,gagliardi2019two,zhu2019novel}. In recent years, researchers have proposed several modifications in the architecture of the RBF with the intention to achieve better convergence rates. For instance, the architectures proposed in \cite{New2}, the algorithm attains faster error convergence by utilizing fewer number of nodes. The new architecture of RBF proposed in \cite{New4}, has been developed using fuzzy clustering techniques. In \cite{timeRBF}, an intelligent adaptive learning method for RBFNN is proposed in which the calculation of the derivative is circumvented so that the algorithm converges at a faster rate.  Verification of the method is carried out using simulations for tracking control and non-linear system identification. Several researchers employed RBF for task oriented applications, for instance, in \cite{meng2021adaptive}, critical water quality parameters have been predicted during the treatment of waste water. A modular RBFNN is designed in order to deal with the complex problems encountered by the human brain in daily life \cite{qiao2020novel}, this approach is inherited from neurophysiology and neuroscience. An improved version of RBF has proved its efficacy in bank credit risk management using optimal segmentation \cite{li2020application}. Several researches have been done for the localization of various types of vehicles, for instance, 5G assisted unmanned aerial vehicle is localized by the help of RBF neural network \cite{annepu2021radial}. An adaptive kernel has been proposed in \cite{khan2016novel} for the RBF neural network. This algorithm makes use of both the Euclidean and cosine distances which are measured in an adaptive manner. This fusion consequently leads to faster error convergence. The algorithm has been evaluated on several signal processing case studies and resulted in superior performance gain compared to the conventional approaches. A fractional gradient descent based RBFNN has been proposed in \cite{FRBF}, in which the weight update rule has been derived using fractional gradient methods. The method compared to the conventional approach  achieves improved results on the signal processing problems including pattern classification, function approximation, time series prediction and non-linear system identification.

In order to achieve accelerated convergence, in the recent past a variety of techniques have been developed. One such technique is called the non-uniform update gain \cite{ummatov2019adaptive}, the use of which has been suggested for the rapid convergence rate in linear filters \cite{givens2009enhanced, haweel1992class, evans1993analysis, harris1986variable}. This is achieved by replacing the constant update gain matrix with the non-uniform update matrix containing the corresponding values for multiplication with the input coefficient. This implication has shown significant improvement in obtaining faster convergence. A similar idea has been extended for the non-linear filters \cite{usman2019quantum}, in which the non-uniform update gain was utilized for the second order expansion of Volterra series. The non-uniform gain for the input coefficients was obtained by exploiting the inherent nature of Jackson's derivative from the $q$-calculus. In \cite{givens2009enhanced} it is suggested that a non-uniform learning rate can accelerate the convergence, since the $q$-gradient descent has non-uniform q factor, which upon multiplication results in non-uniform learning rate, therefore, we can expect better convergence with q-gradient descent \cite{qLMS}.

The concept of q-calculus is known as the calculus with no limits and this approach has been greatly utilized by a number of researchers and has gained overwhelming success in fields ranging from quantum theory, mathematics, and physics \cite{ernst2000history}. Jackson established the concepts of q-derivative \cite{jackson1909xi}, and q-integral \cite{jackson1910q}, the notions since then have been utilized towards the development of several stochastic gradient algorithms \cite{qLMS_WF,qLMS}. Contrary to the conventional gradient which computes the tangent, the q-derivative evaluates the secant of the function, consequently taking larger steps towards the optimal solution resulting in faster convergence.

Additionally, substantial work has been carried out to avert the constant learning rate of the gradient algorithms \cite{VSS, messalti2017new,VPFLMS, RVSS1, RVSSFLMS, hussain2019improved, sadiq2020q}. The dynamic updates in the learning rate has been shown to effectively contribute towards achieving faster convergence rates. A notable method is proposed in \cite{EqLMS}, which uses the notions of non-uniform update gain along with the implementation of variable step size by utilizing the concepts of signal normalization and error correlation energy. 

Considering merits of variable step size and $q$-gradient subsequently leading to the non-uniform gains of the coefficient in the adaptive filter, in this research, we propose to extend these concepts for the RBF neural network. In particular, the following contributions have been made in this research work:
\begin{enumerate}
	\item We propose a novel learning algorithm for the RBFNN using a modified stochastic gradient descent algorithm based on the notion of q-gradient.  
	
	\item A thorough mathematical analysis for the steady state performance of the proposed algorithm is presented which validates the simulation results of the experiments conducted for the system identification problem. 
	
	\item An optimal solution (i.e the Wiener solution) and convergence analysis is presented for the proposed method and its stability bounds are derived.
	
	\item Analytical results are validated through computer simulations for the sensitivity, transient and steady state behavior of the proposed $q$-RBFNN.
	
	\item Extensive comparative analysis of the proposed work is carried out with the contemporary approaches.

	\item  Additionally, an adaptive framework is designed, which aims to achieve higher convergence rate without compromising the steady-state error.
\end{enumerate}

This research is intended to be a  fusion of q-calculus with the RBF neural network to develop an advancement in the inherited gradient descent optimization. Sophisticated stochastic methods of learning developed for the optimization of free parameters in the RBF algorithm \cite{RBFTr8,BeeRBF,jafrasteh2017hybrid}, are not the focus of this research work.

The rest of the paper is organized as follows: The proposed learning approach, along with the mathematical analysis of the proposed method, optimal solution and the convergence and sensitivity analysis of the proposed method has been discussed in Section \ref{sec:1}. The time varying $q$-RBF method and its robustness is discussed in Section \ref{sec:tvqrbf}. The performance evaluation of the proposed method and the comparative analysis has been carried out in Section  \ref{sec:nsi}, and the paper is finally concluded in Section  \ref{conclusion}.

\section{The Proposed \emph{q}-gradient-based RBF Neural Network (\emph{q}-RBFNN)} \label{sec:1}

\begin{figure}[!ht]
	\begin{center}
		\centering 
		\includegraphics[width=10cm]{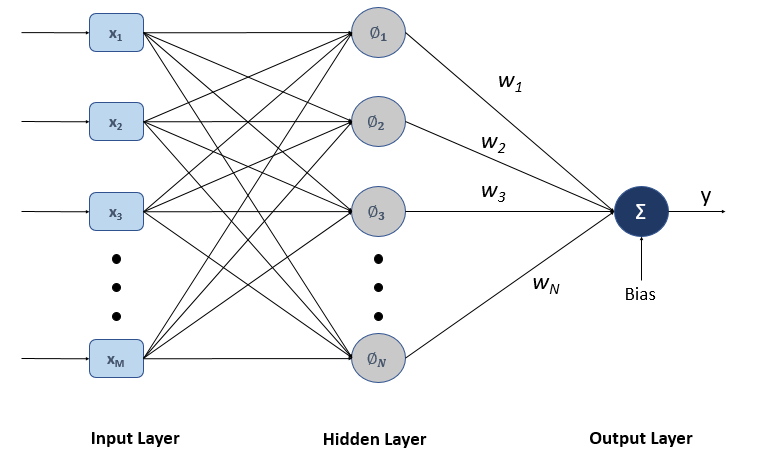} 
	\end{center}
	\caption{RBF neural network (RBFNN) Architecture.}
	\label{rbfarch}
\end{figure}

A typical RBFNN as shown in Fig.\ref{rbfarch}, is a three layered architecture comprised of (1) an input layer, (2) a non-linear hidden layer and (3) an output layer.

To understand the mapping of of the RBF, consider an input vector $\mathbf{x} \in \mathbb{R}^{M}$,  $s:\mathbb{R}^{M}\rightarrow\mathbb{R}^{1}$, is given as

\begin{eqnarray}
y=\sum_{i=1}^{N}w_i\phi_i(\left\|\mathbf{x}-\mathbf{c}_i\right\|)+b,
\end{eqnarray}

where $\mathbf{c}_i \in \mathbb{R}^{M}$ depicts the center locations of the RBF network, the connection between the hidden and the output layer is weighted $\mathbf{w}_i$, and a bias $b$ is added to the output layer. The number of neurons in the hidden layer are $N$ and $\phi_i$ is the basis of each neuron in the hidden layer. The number of output neurons are equivalent to the number of classes, here we consider a single output neuron. There are several choices of kernels among which gaussian, multiquadratics and inverse multiquadratics are often used for RBF networks \cite{fifth}, however, gaussian kernel as shown in \eqref{eq: gaussian_kernel}, due to its versatility is mostly  employed \cite{seventeen}.
\begin{eqnarray} \label{eq: gaussian_kernel}
\phi_i(\left\|\mathbf{x}-\mathbf{c}_i\right\|)=\exp\left(\frac{-\left\|\mathbf{x}-\mathbf{c}_i\right\|^2}{\sigma^{2}}\right),
\end{eqnarray}   

Here $\sigma$ represents the Gaussian kernel's spread. The kernels serve the purpose of realizing the distance from the center of the network. The commonly used distance is the Euclidean distance, although other distance metrics such as cosine distance has been suggested to have complimentary properties in comparison with the Euclidean distance metric \cite{drmoin}.

\begin{eqnarray}
\label{gamma1}
\phi_{i1}(\mathbf{x}.\mathbf{c}_i)=\frac{\mathbf{x}.\mathbf{c}_i}{\left\|\mathbf{x}\right\|\left\|\mathbf{c}_i\right\| + \gamma},
\end{eqnarray}
in exceptional cases where the denominator terms $\left\|\mathbf{x}\right\|$ or $\left\|\mathbf{c}_i\right\|$ may become zero, the equation (\ref{gamma1}) will be indeterminate. To avoid such situation, a small constant term $\gamma > 0$ is added.

\subsection{Proposed \emph{q}-RBF learning approach}

The concepts of q-calculus has been successfully utilized in variety of fields including signal processing, mathematics and quantum theory \cite{qbook2,qbook3,qbook4,qbook5}.

In \cite{kac2001quantum}, a method of obtaining the function's derivative is described as follows:

\begin{equation}
d_{q_i}(g(k)) = g(q_i k)-g(k).
\end{equation}

Differentiating the above equation leads to the result as follows:

\begin{equation}
D_{q_i}(g(k)) = \frac{d_{q_i}(g(k))}{d_{q_i}(k)} = \frac{g(q_i k)-f(k)}{(q_i -1)k}.
\end{equation}

In the above equation if $q\rightarrow1$ is substituted, the q-derivative function is similar to the classic derivative.

Here we introduce the proposed learning approach of RBFNN based on the q-gradient descent algorithm.
To derive the weight update rule, consider the RBFNN architecture shown in Fig \ref{rbfarch}, the output of which at the $n$th iteration can be written as:
\begin{eqnarray}
\label{map}
y(n)=\sum_{i=1}^{N}w_i(n)\phi_i(\mathbf{x},\mathbf{c}_i)+b(n),
\end{eqnarray} 

The number of neurons in the hidden layer are $M$ which formulates the final result. The values of synaptic weights  $w_i(n)$ along with the bias $b(n)$ are adapted at each iteration. The cost function $\mathcal{E}(n)$ is the instantaneous error $e(n)$, found by taking difference of actual and the desired output as shown in \eqref{eq: cost}

\begin{equation}
\label{eq: cost}
\mathcal{E}(n)=\frac{1}{2}(d(n)-y(n))^{2} = \frac{1}{2}  e^2(n),
\end{equation}

The weight update equation is derived from the conventional gradient descent method:
q-gradient is incorporated in the conventional gradient descent method to obtain the weight update rule as shown in \eqref{eq:Weight}

\begin{equation}\label{eq:Weight}
w_{i}(n + 1) = w_{i}(n) - \mu \nabla_{q_{i},w_{i}} \mathcal{E}(n),
\end{equation}

The parameter $q_i$ controls the gradient and $\mu$ denotes the step size.
Evaluating the factor $-\nabla_{q_{i},w_{i}}\mathcal{E}(n)$ for $N=i=1$ by: 

\begin{equation}\label{Normal} -\nabla_{q_{i},w_{i}} \mathcal{E}(n) =  - \frac{d_{q_{i}} \mathcal{E}(n)}{d_{q_{i}} e(n)} \times\frac{d_{q_{i}} e(n)}{d_{q_{i}} y(n)} \times \frac{d_{q_{i}} y(n)}{d_{q_{i}} w_{i}(n)} 
\end{equation}

simplifying the partial derivatives in (\ref{Normal}) yields:
\begin{equation}\label{Normal_Sim}
- \nabla_{q_{i},w_{i}} \mathcal{E}(n) =  \frac{(q_{i}+1)}{2} \phi_i(\mathbf{x},\mathbf{c}_i) e(n),
\end{equation}

using \eqref{Normal_Sim} equation \eqref{eq:Weight} is reduced to be:		
\begin{equation}\label{weightf}
w_{i}(n + 1) = w_{i}(n) + \mu \; \frac{(q_{i}+1)}{2} \; \phi_{i}(\mathbf{x},\mathbf{c}_{i}) e(n) .	
\end{equation}

Similarly, $b(n)$ can be updated by:
\begin{equation}
b(n + 1) = b(n) +  \mu \; \frac{(q_{0}+1)}{2} \; e(n) .
\end{equation}

Extending it in a similar fashion for $N$, the controlling parameters $q_i$ in \eqref{weightf} can be contained in the diagonal matrix $\mathbf{G}$ as shown in \eqref{G}. 
\begin{eqnarray}\label{G}
	{\rm diag}(\mathbf{G}) = [\frac{(q_{1}+1)}{2}, \frac{(q_{2}+1)}{2},.....\frac{(q_{N}+1)}{2}]^{\intercal}.
\end{eqnarray}

Hence, the weight vector $\mathbf{w}$ in \eqref{weightf}, can be written as:
\begin{equation}\label{weightg}
\mathbf{w}(n + 1) = \mathbf{w}(n) + \mu \;  \mathbf{G} \; \mathbf{\phi}(\mathbf{x},\mathbf{c}) e(n).
\end{equation}	

\subsection{Mathematical Analysis of q-RBFNN}

The optimal solution determines the fact that a stochastic learning algorithm shall converge. To find the optimal solution we carried out theoretical analysis of the proposed method. Consider a scenario of system identification where the  desired output $d(n)$ is shown in \eqref{desired}:

\begin{equation}\label{desired}
d(n)=\sum_{i=1}^{N}\hat{w}_i(n)\phi_i(\mathbf{x},\mathbf{c}_i)+\zeta(n),
\end{equation}

The system to be identified is denoted by $\hat{w}$, having an input node vector $ \textbf{x}(n) $, and $\zeta(n)$ is the additive white noise with zero mean. 
The unknown system $\hat{\textbf{w}}$ is estimated by utilizing the relation in \eqref{map}.

\subsubsection{Optimal Wiener Solution}
We derive the cost function by substituting the error relation in the simple error i.e. $e(n) = d(n) - y(n)$, and the output in eq. \eqref{map}, in the conventional cost function which is based on the mean square error, which is $\mathcal{E}(n) = \frac{1}{2}E[e(n)e^\intercal(n)] = \frac{1}{2}E[|e(n)|^2]$. This yields to the following relation:

\begin{multline}\label{eq:General_Cost_Function}
\mathcal{E}(n) = \sigma_d^2 - \sum\limits_{k=0}^{N} w_k^\intercal(n) p(-k) - \sum\limits_{k = 0}^{N} w_k(n) p^\intercal(-k) \\ + \sum\limits_{k = 0}^{N} \sum\limits_{l = 0}^{N} w_k^\intercal(n) \hat{w}_i(n) r(l - k),
\end{multline}

where $ \sigma_\mathbf{d}^2 \triangleq E[|d(n)|^2] $ is the desired signal power, the cross-correlation between the desired output and the kernel output is $ p(-k) \triangleq E[\phi_i(\mathbf{x},\mathbf{c}_i)d^\intercal(n)]$, and the auto-correlation matrix of the kernel is $ r(l - k) \triangleq E[\phi(\mathbf{x}(n-k),\mathbf{c})\phi(\mathbf{x}(n-l),\mathbf{c}_i)^{\intercal}] $.

The cost function $q$-gradient can be written as:
\begin{multline}
     \nabla_{q_{k},w_k} \mathcal{E}(n) = \frac{d_{q_{k}} \mathcal{E}(n)}{d_{q_k} w_k} = - \frac{(q_{k} +1)}{2} p(-k)  \\ +  \sum\limits_{l = 0}^{N} \frac{(q_{k} +1)}{2} \hat{w}_i(n) r(l - k). 
\end{multline}\label{eq:Wiener2}


To achieve an optimal solution, $ \nabla_{q_{k},w_k} \mathcal{E}(n) $ is set to zero:
\begin{equation}
p(-k) = \sum\limits_{i = 0}^{N} \hat{w}_i(n) r(l - k),
\end{equation}
and
\begin{equation}
\mathbf{\hat{w}} = \mathbf{r}^{-1}\mathbf{p}.
\end{equation}

A least square problem's closed form of Wiener solution is mentioned in the above equation. It is of great interest that the optimal solution of the proposed algorithm gives the Wiener solution without any additional parameter(s). The minimum square error at $\mathbf{\hat{w}}$ is also same as optimal Wiener power.
\begin{eqnarray}\label{q_gradient2}
\boldsymbol{\xi}_{min} = E[\zeta^{2}(n)],
\end{eqnarray}

where $\zeta$ is the Gaussian noise.  This outcome shows that the proposed method shall cease the weight update upon attaining the Weiner solution and guarantees the convergence of algorithm.

\subsubsection{Convergence Analysis}
The convergence analysis is performed for the mean error performance of the proposed algorithm with common assumptions \cite{W1, Wiener_existence}: The noise has a Gaussian distribution with zero mean and unit variance, the input vector $\mathbf{x}$ is independent and identically distributed i.i.d and the such parameters of kernel function $\mathbf{\phi}$  are  selected that signals remain linearly separable in the Kernel space after mapping.

The weight error vector is defined as $\Delta_{w}(n)=\boldsymbol{w}(n)-\boldsymbol{\hat{w}}$, 

\begin{equation}\label{error}
e(n)=\Delta_{w}^{\intercal}(n)\phi(\mathbf{x}(n),\mathbf{c}) + \zeta(n)
\end{equation}

After substituting $\boldsymbol{w}(n)=\Delta_{w}(n)+\boldsymbol{\hat{w}}$, and $e(n)$ in \eqref{weightg}
, we get

\begin{multline}\label{Mean_analysis1}
   \Delta_{w}(n+1) = \Delta_{w}(n) +\eta \boldsymbol{G} \phi(\mathbf{x}(n),\mathbf{c}) \\ \Delta_{w}(n)^{\intercal}(\phi(\mathbf{x}(n),\mathbf{c})  + \zeta(n) 
\end{multline}


After simplification it results in 
\begin{eqnarray}\label{Mean_analysis2}
E\left[\Delta_{w}(n+1)\right] = \left(\mathbf{I} - \mu \mathbf{A}\right) E\left[\Delta_{w}(n)\right].
\end{eqnarray}
and upon further simplification gives
\begin{eqnarray}\label{Mean_analysis3}
E\left[\Delta_{w}(n)\right] = \left(\mathbf{I} - \mu \mathbf{A}\right)^{n} E\left[\Delta_{w}(0)\right].
\end{eqnarray}
where $\mathbf{A} = \boldsymbol{G} E\left[\phi(\mathbf{x}(n),\mathbf{c})( \phi(\mathbf{x}(n),\mathbf{c})^{\intercal}\right]$ and $E[]$ is the expectation operator.

For convergence 
\begin{equation}
0 < \mu < \frac{1}{max\{(q_1 +1)\lambda_{1},\dots,(q_N +1)\lambda_{N} \}}
\end{equation}
In case when all $q_i$'s are equal to $q$
\begin{equation}
0 < \mu < \frac{1}{(q+1)\lambda_{max}}
\end{equation}
where $\lambda_{max}$ is the maximum eigenvalue of $\mathbf{r}$ and it implies that  
\begin{equation}
\mu_{\max} = \frac{1}{||\phi(\mathbf{x}(n),\mathbf{c})||_{G}^{2}}
\end{equation}

\subsection{Sensitivity Analysis of the proposed \emph{q}-RBF NN} \label{sec:sensitivity_analysis}
In this experiment, we analyze the sensitivity of the $q$-RBF algorithm with respect to the parameter $q$.  In particular, we choose a pattern classification problem and compare the Mean Square Error (MSE) learning curves of the proposed $q$-RBF algorithm for different values of $q$-parameter.  Simulation parameters used are as follows: for the proposed $q$-RBF algorithm, we investigated four different values of $q$ which are $q=2$, $q=5$, $q=10$, and $q=12$.  The task is to cluster $100$ random signals obtained from two Gaussian distributions of means $0.25$ and $0.75$ and variance $0.1$ into two different groups namely class $0$ and class $1$.  Two Gaussian kernel neurons are used with spread of $0.1$ and center values are obtained using k-means clustering.  The output class signal is disturbed by adding a Gaussian noise of signal-to-noise ratio (SNR) of $20$ dB.  The model was trained on $200$ epochs with the learning rate value chosen to be $0.10$. The simulations are repeated $100$ times and mean results are reported.

\begin{figure}[!ht]
		\centering{
		\includegraphics[width=8cm]{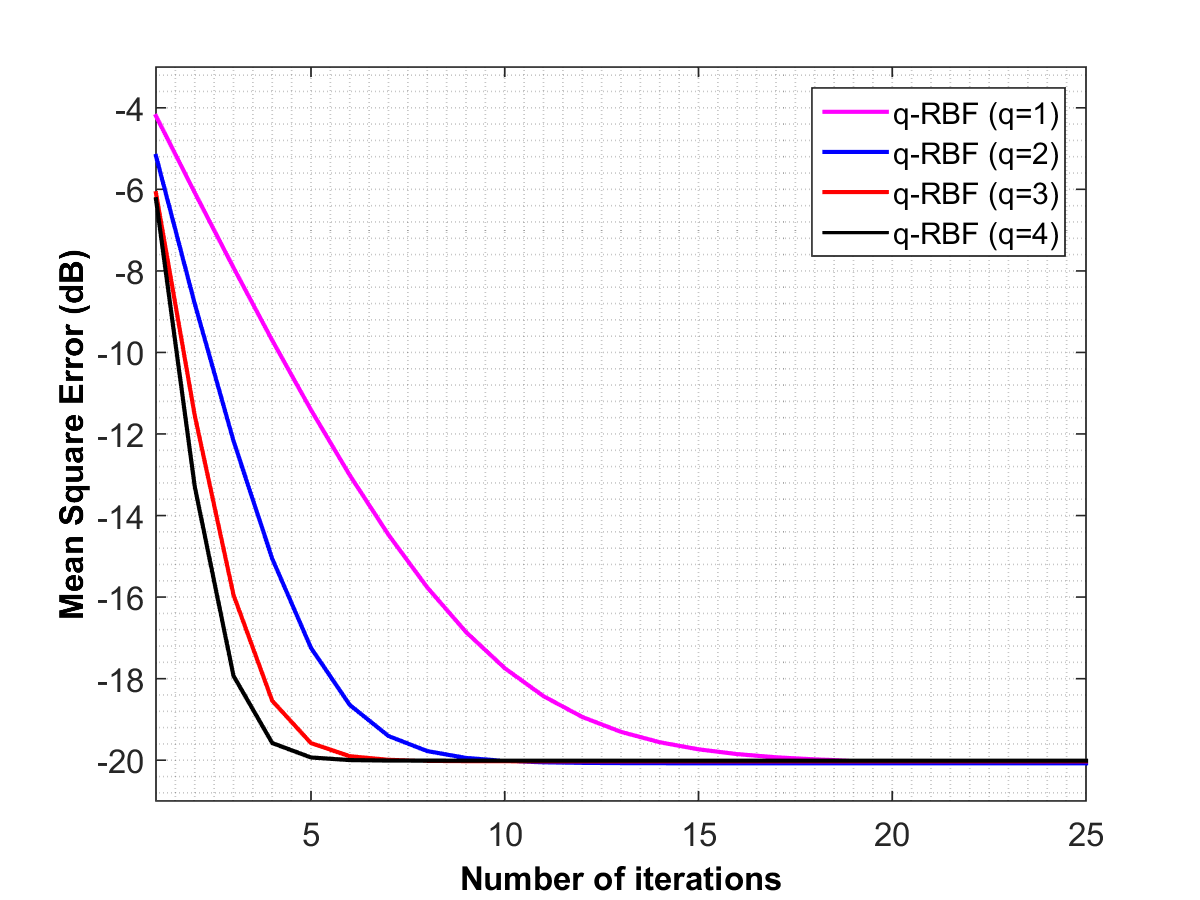}
		\hspace{0.1cm}\includegraphics[width=8cm]{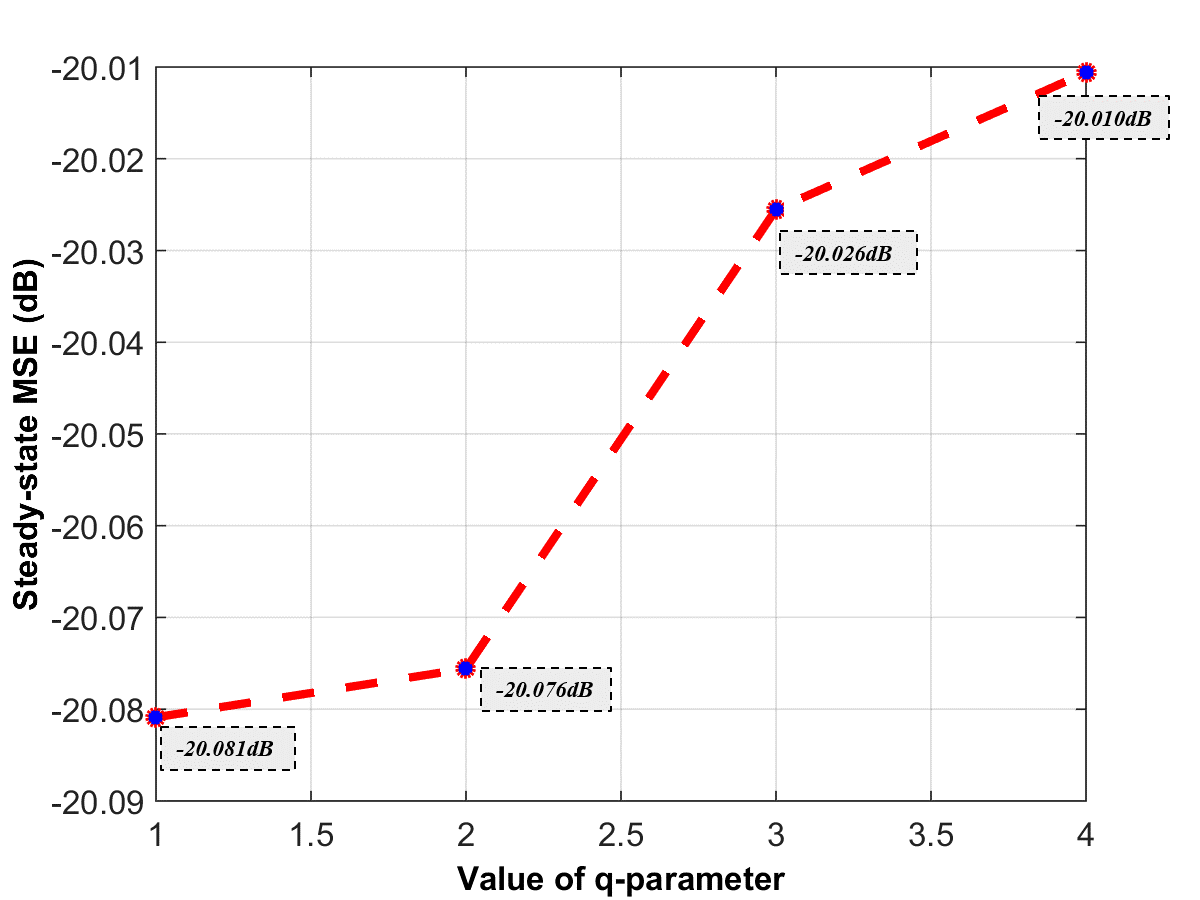}}
	\caption{Sensitivity analysis of the proposed algorithm}
	\label{fig:sensitivity1}
\end{figure}

Fig. \ref{fig:sensitivity1}(a) clearly shows that for higher values of $q$, the proposed $q$-RBF algorithm exhibits faster convergence similar to that of classical least mean square (LMS) algorithm.  The final steady-state error in all cases is close to the disturbance in system i.e., $20$ dB.  However, the faster convergence is achieved at a cost of steady-state error Fig. \ref{fig:sensitivity1}(b), shows the increase in residual error vs increment in value of $q$.  This behaviour motivated us to device a mechanism for time-varying $q$-parameter which can provide highest convergence with lowest steady-state error.  The details of the proposed time-varying $q$-RBF is given in section \ref{sec:tvqrbf}.

\subsection{Validation of Mathematical Analysis using simulation}
\begin{figure}[!ht]
	\begin{center}
		\centering
		\includegraphics[ height=6cm, width=8cm]
		{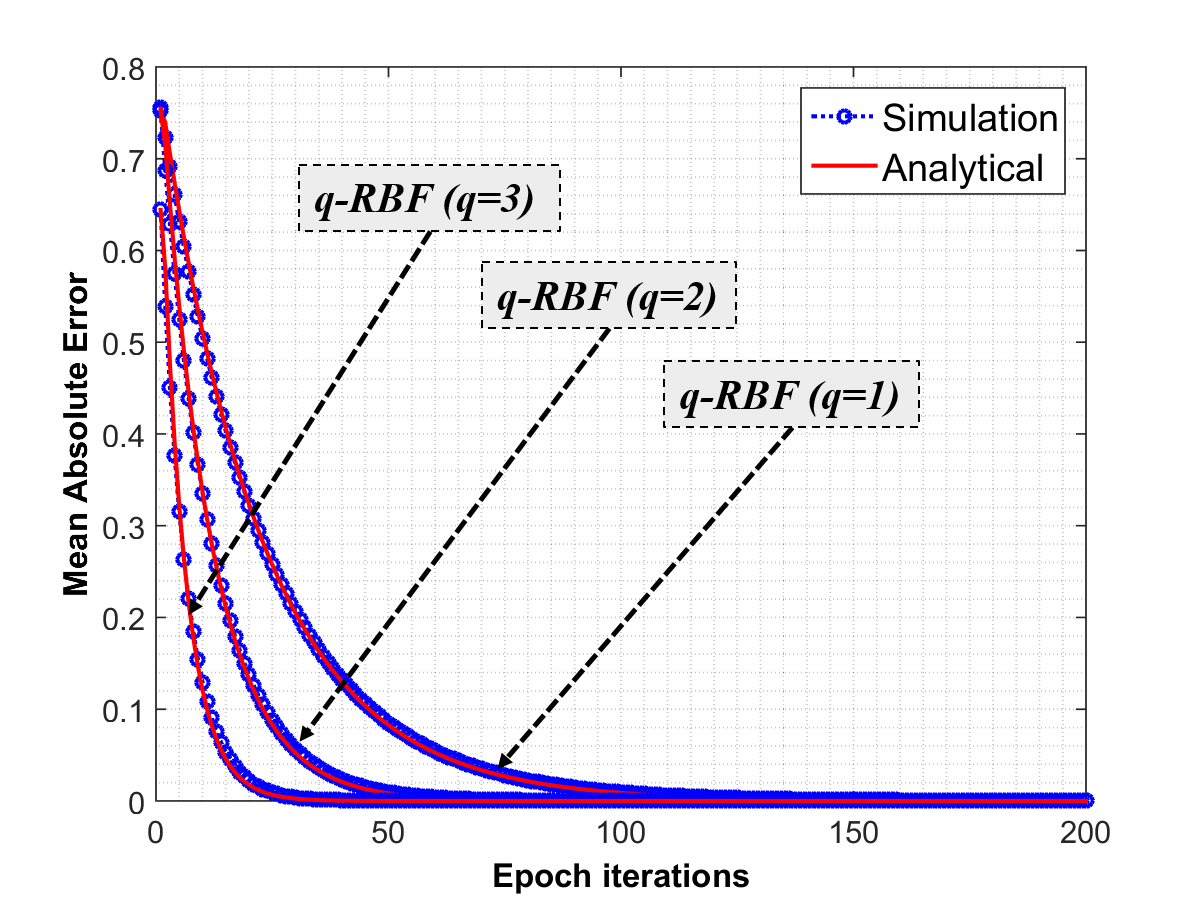}
	\end{center}
	\caption{Mean convergence of the proposed algorithm on different values of $q$}
	\label{fig:sensitivity}
\end{figure}

The same experiment of section \ref{sec:sensitivity_analysis} was carried out again expect this time no disturbance signal is added and instead of MSE, mean absolute error (MAE)  between optimal weights $\mathbf{\hat{w}}$ and estimated weights $\mathbf{w}$ is calculated. Analysis results are obtained using eq. \eqref{Mean_analysis3}.  The simulations are performed $100$ times are mean results of MAE are reported for three different values of $q$, i.e., ($1$, $2$, and $4$).  Fig. \ref{fig:sensitivity} shows the MAE curves for simulation and analysis.  The analysis results are well matched with simulation results and the mean correlation coefficient of analysis and simulation MAE values is almost $1$ i.e., $0.9996$

\section{Design of Time-varying \emph{q}-RBF}
\label{sec:tvqrbf}
From sensitivity analysis in section \ref{sec:sensitivity_analysis}, it is evident that selection of an appropriate $q$ value contributes toward improvement in the convergence performance of the algorithm. The convergence rate can be further increased for $q > 1$, therefore, a time-varying $q$-RBFNN design is proposed.

\subsection{Analysis of the Robustness of \emph{q}-RBFNN using Small Gain Theorem}
\begin{figure}[!ht]
	\begin{center}
		\centering
		\includegraphics[ height=4cm, width=9cm]
		{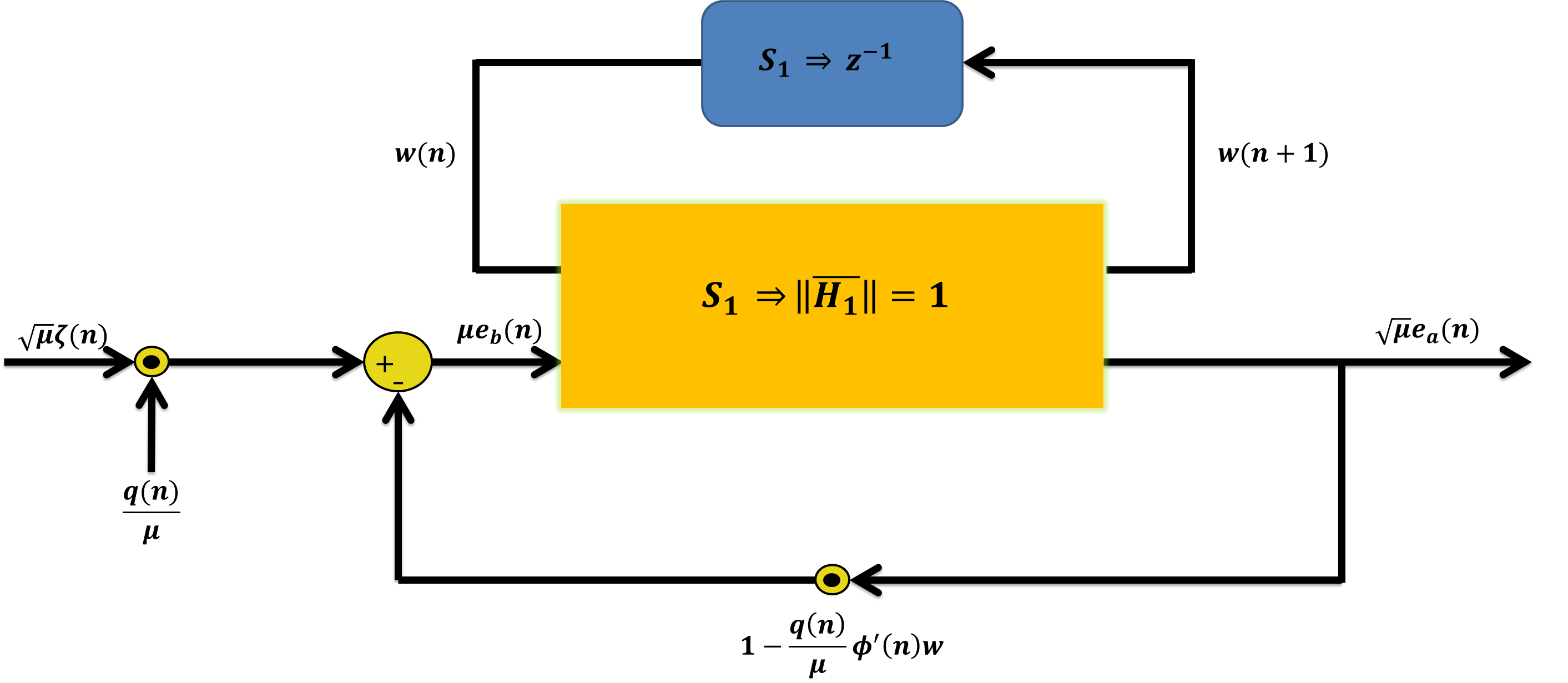}
	\end{center}
	\caption{Lossless Mapping of a closed loop system for $q$-RBFNN algorithm}
	\label{fig:sgt}
\end{figure}

The small gain theorem \cite{teel1996nonlinear} serves the purpose very well in order to discuss the robustness of the equation \eqref{error}. The theorem can be explained by the relation $||S_1|| * ||S_2|| < 1$ where $S_1$ and $S_2$ are considered to be two stable systems which are connected together in a closed loop as shown in Fig. \ref{fig:sgt}. $S_1$ is said to be a feed-forward block and $S_2$ is considered to be a feedback block. To ensure the convergence rates and reliable training scheme, the possibilities will be derived for the learning rate $q$. The small gain theorem  provides sufficient conditions to the stability of finite-gain $l_2$ for the efficient mapping of noisy input signal with the estimated error sequence.

A robust algorithm has consistent estimation error under the influence of perturbations regardless of their nature. This property is of great importance in situations where the prior statistical knowledge is not present. The robustness therefore, implies a positive constant as an upper bound on the estimation error to the perturbation energy as shown in \eqref{robust} .
\begin{equation}\label{robust}
    \frac{Estimation Error}{Noise} \leq 1
\end{equation}

Next, while adapting the weights from the $nth$ iteration to the $(n+1)th$ iteration, we have to perform a lossless mapping in between the estimation errors for all the time instances $n$. The transformation from $x$ to $y$ as $y =H(x)$ in accordance with the relation i.e. $||H(x)||^2 \leq||x||^2$ is considered to be a lossless mapping assuring that the output energy will always be lesser than the input energy. We define the disturbance error for the analysis i.e. $\tilde{\zeta}(n)=e(n)-e_a(n)$ To explain the lossless mapping between estimation errors $e_a(n)$ and $e_b(n)$, a feedback system is considered. Where, $e_a(n)$ is apriori error and $e_b(n)$ is considered as a posterior error which can be defined as in eq \eqref{ea} and \eqref{eb}.

\begin{equation}\label{ea}
 e_a(n) =\phi(n)\mathbf{\tilde{w}}(n) 
 \end{equation}
\begin{equation}\label{eb}
    e_b(n) =\phi(n)\mathbf{\tilde{w}}(n+1) 
\end{equation} 
Where, $\mathbf{\tilde{w}}$ is known as weight error vector distinguishing the optimal and its estimated weight as $\mathbf{\tilde{w}} = w_o - w(n)$. Therefore, eq \eqref{ea} and eq \eqref{eb} can be rewritten as eq \eqref{ea2} and eq \eqref{eb22}, acquired using eq \eqref{weightf}, respectively.

\begin{equation}
\label{ea2}
 e_a(n) =\phi(n)w_o(n) - \phi(n)w(n)   
\end{equation}
\begin{equation}
    \label{eb22}
    \begin{split}
        e_b(n) &= \phi(n)[\mathbf{\tilde{w}}(n) - \mu q(n)\phi^T(n)e(n)]\\
        &=e_a(n)-\mu q(n)||\phi(n)||^2e(n)
    \end{split}
\end{equation}
Hence, we define posterior error $e_b(n)$ in the form of $\mu$ in eq \eqref{eb1} and \eqref{eb2}.
\begin{equation}\label{eb1}
    e_b(n) =e_a(n)-\mu q(n)||\phi(n)||^2e(n)
\end{equation}
\begin{equation}\label{eb2}
    q(n)e(n)=\mu\{e_a(n)-e_b(n)\}
\end{equation}
The weight update rule mentioned in eq \eqref{weightg} can be expressed in error recursion form as:
\begin{equation}\label{wer}
    w(n+1)=w(n)-\mu q(n)\phi^T(n)\{e_a(n)-e_b(n)\}.
\end{equation}
To calculate the energy taking $l_{2}$ norm of \eqref{wer}:
\begin{multline}\label{wer2}
    ||w(n+1)||^{2}  = ||w(n)||^{2}- 2[w(n) \mu{q}(n) \phi^{T}(n)\{e_{a}-e_{b}\}]+ \\\mu^{2} [{q}(n) \phi^{T}(n)\{e_{a}-e_{b}\} {q}(n) \phi^{T}(n)\{e_{a}-e_{b}\} ] 
\end{multline}

Given that $\mu$ is a small value, therefore, we can ignore its higher power, resulting in:
\begin{equation}
||w(n+1)||^{2}  = ||w(n)||^{2}- 2[w(n) \mu{q}(n) \phi^{T}(n)\{e_{a}-e_{b}\}]
\end{equation}

\begin{equation}\label{}
    ||w(n+1)||^2+2\mu q(n)e_a(n) = ||w(n)||^2+2\mu q(n)e_b(n)
\end{equation}

\begin{equation}\label{wnew}
    \frac{||w(n+1)||^2+2\mu q(n)e_a(n)}{||w(n)||^2+2\mu q(n)e_b(n)}=1
\end{equation}
The above form \eqref{wnew} is valid for all the possible learning rates.

By using the relations in \eqref{error}, \eqref{ea}, \eqref{eb1}, and \eqref{eb2} can be expressed as:
\begin{equation}
    \begin{split}
        e_b(n)&=e_a(n)-\frac{q(n)}{\mu}\{e_a(n)+\zeta(n)\},\\
        e_b(n)&=\left[1-\frac{q(n)}{\mu}\right]e_a(n)-\frac{q(n)}{\mu}\zeta(n),\\
        -\sqrt{\mu}e_b(n)&=\frac{q(n)}{\sqrt{\mu}}\zeta(n)-\left[1-\frac{q(n)}{\mu}\right]\sqrt{\mu}e_a(n).
    \end{split}
\end{equation}
From the above expression, it is evident that the complete mapping from the actual disturbances $\sqrt{\mu}\zeta(n)$ to the resulting estimation errors $\sqrt{\mu}e_a(n)$ can be represented by the closed loop system shown in the Fig. \ref{fig:sgt}. For our case, the small gain theorem can be expressed as:
\begin{equation}
    \Delta(M)=\max_{0 \leq n \leq M}\bigg|1-\frac{q(n)}{\mu}\bigg|.
\end{equation}
From the above equation, $\Delta(M)$ is said to be an absolute maximum gain of the closed loop for time period of $0 \leq n \leq M$. For a system shown in \ref{fig:sgt}, the small gain theorem states the guaranteed stability if the multiplication results of norm of feedforward  and feedback mappings are bounded by 1. For the given mapping the norm of the feedforward block is equal to 1, and since the the feedback norm $\Delta(M) < 1$  the overall stability requirement is guaranteed. There must be a specific range of the learning rate which can be defined as:

\begin{equation}
\label{eq:qmax}
    0 < q(n) < \frac{1}{\mu ||\phi(n)||^2}
\end{equation}

\subsection{Proposed design for adaptive \emph{q}}
\label{sec:adaptive_qrbf}
We propose the following time varying rule for the $q$ parameter: 
\begin{equation}\label{q1}
    q(n+1) = \beta q(n) +\gamma e^{2}(n)
\end{equation}
where, $0 < \beta < 1$, and $\gamma > 0$ with
	\begin{equation}\label{q2}
	q(n+1) = \left\{ \begin{array}{rcl}
	q_{max} \ for & q(n+1) > q_{max}\\ 
	q(n+1),  & otherwise\\ 
	\end{array}\right.
	\end{equation}
Here, $q_{max}$ is calculated using \eqref{eq:qmax}, the parameter $\beta$ is a positive value $0 < \beta < 1$, and is dependent on its own past value, while the constant $\gamma > 0$. The adaptation rule in the mappings \eqref{q1} and \eqref{q2} suggests that greater error correlation in the initial stage would prompt a large learning rate and shall reduce as the system approaches to the steady state. This desired behavior is similar to that proposed for the standard LMS algorithm and its variants \cite{EqLMS, qLMS_WF,qLMS}. 

\section{Nonlinear System Identification using Proposed \emph{q}-RBF NN}
\label{sec:nsi}

\begin{figure}[!ht]
	\centering
	\centerline{\includegraphics[width=8cm]{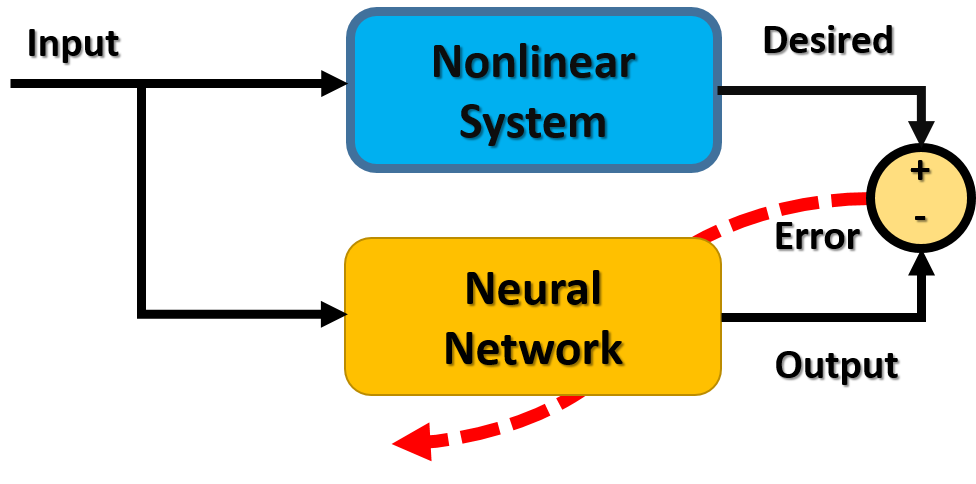}}
	
	\caption{System identification block diagram}
	\label{plant1}
\end{figure}

The linear models are employed due to their easy implementation and assurance of robustness. However, many practical and industrial applications demand highly complex and nonlinear system modeling. To this end, neural networks due to their universal approximation capabilities comes in handy. They present efficient modeling solution by utilizing the system's inputs and outputs. One of such system has been shown in Fig \ref{plant1}. To determine the efficacy of the proposed method, we evaluate its performance for three system identification problems. Furthermore, the effectiveness of the proposed algorithm is tested for the prediction of chaotic time series which inherit high degree of complexity \cite{sadiq2018chaotic}.

\subsection{Highly nonlinear control system identification}
\label{secHNI}   
A non-linear system is defined by the following transfer function:

\begin{multline}
    \label{equ:system}
y(t)=a_1r(t)+a_2r(t-1)+a_3r(t-2) \\ +a_4[\cos(a_5 r(t))+e^{-|r(t)|}]+n(t)
\end{multline}


The output of the system is $y(t)$, the input is $r(t)$ with polynomials $a_i$s representing the zeros of the system. Furthermore, random noise $n(t)$ is added to the input with the characteristics $\mathcal{N}(0,\sigma_d^2)$. For training of the model a rectangular signal with additive Gaussian noise of $-10dB$ has been used. One period of the training signal has $500$ samples with first $250$ samples being set to 1 and the rest set to -1. Two periods of training signal, i.e. total signal length of $1000$ was used to train the model.

Testing is carried on a rectangular pulse signal with $-20$ dB Gaussian noise and $2.5\times$ higher frequency than the input and the performance of the proposed method is compared with conventional RBFNN and its fractional design $f$-RBF \cite{FRBF}.


\subsubsection{Architecture and Model Configurations}\label{sec:model}

Three layered  RBFNN structure is considered with three inputs in the input layer, $6$ neurons in the hidden layer and an output layer. The spread ($\sigma$) of the Gaussian kernel is kept $1$ and experiment is performed for $100$ independent runs, the mean of which has been reported. 

The polynomials $a_i$s of the proposed method were chosen to be $2$, $-0.5$, $-0.1$, $-0.7$ and $3$ for $i =$ $1$, $2$, $3$, $4$ and $5$ respectively while $\beta$, $\gamma$ and $q_{\max}$ of time-varying $q$-RBFNN are chosen to be $0.9$, $5$, and $5$ respectively.
The tuning parameters of RBFNN, $f$-RBFNN, and proposed $q$-RBFNN are empirically opted in order to achieve the optimal results during training, and absolute value of gradient of the fractional update term is used to avoid complex values. There are several hyper-parameters in RBFNN and $f$-RBFNN algorithm including the mixing parameter $\alpha$ of gradient in $f$-RBFNN, which  is set to be $0.5$, learning rates $\eta$ and $\eta_\nu$ were both set to $1\times10^{-2}$, the fractional derivative power $\nu=0.9$ is used.

\begin{figure}[!ht]
	\begin{center}
		\centering
		\includegraphics[width=8cm]
		{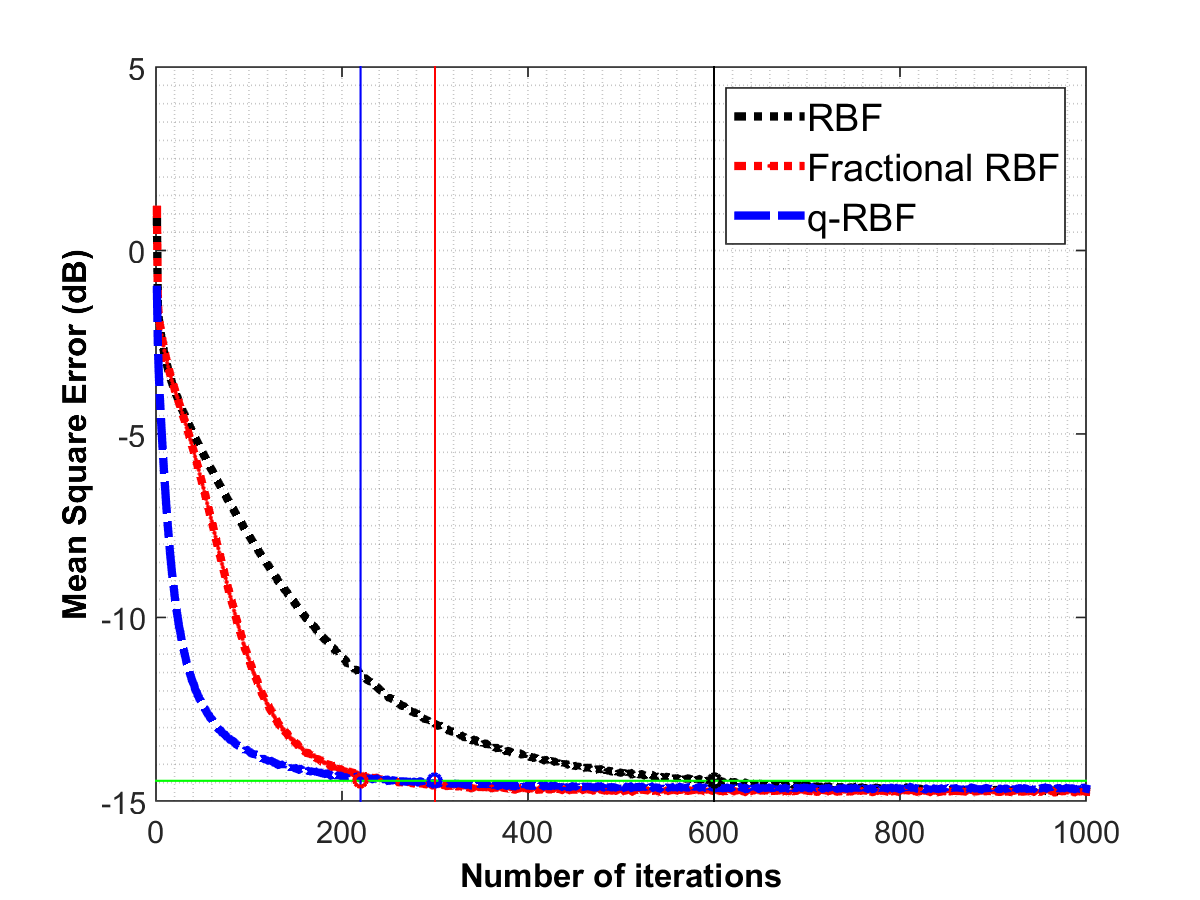}
	\end{center}
	\caption{MSE behavior during training for non-linear system identification problem}
	\label{fig:MSE}
\end{figure}

Fig. \ref{fig:MSE} depicts the MSE curves of the trained algorithms. It can be observed that the proposed  $q$-RBFNN achieves an average MSE of $-14.5$dB in just $220$ iterations, thereby outperforming RBFNN and $f$-RBFNN which required $300$ and $600$ iterations respectively to achieve similar MSE value.

\begin{figure}[!ht]
	\begin{center}
		\centering  
		\includegraphics[width=8cm]
		{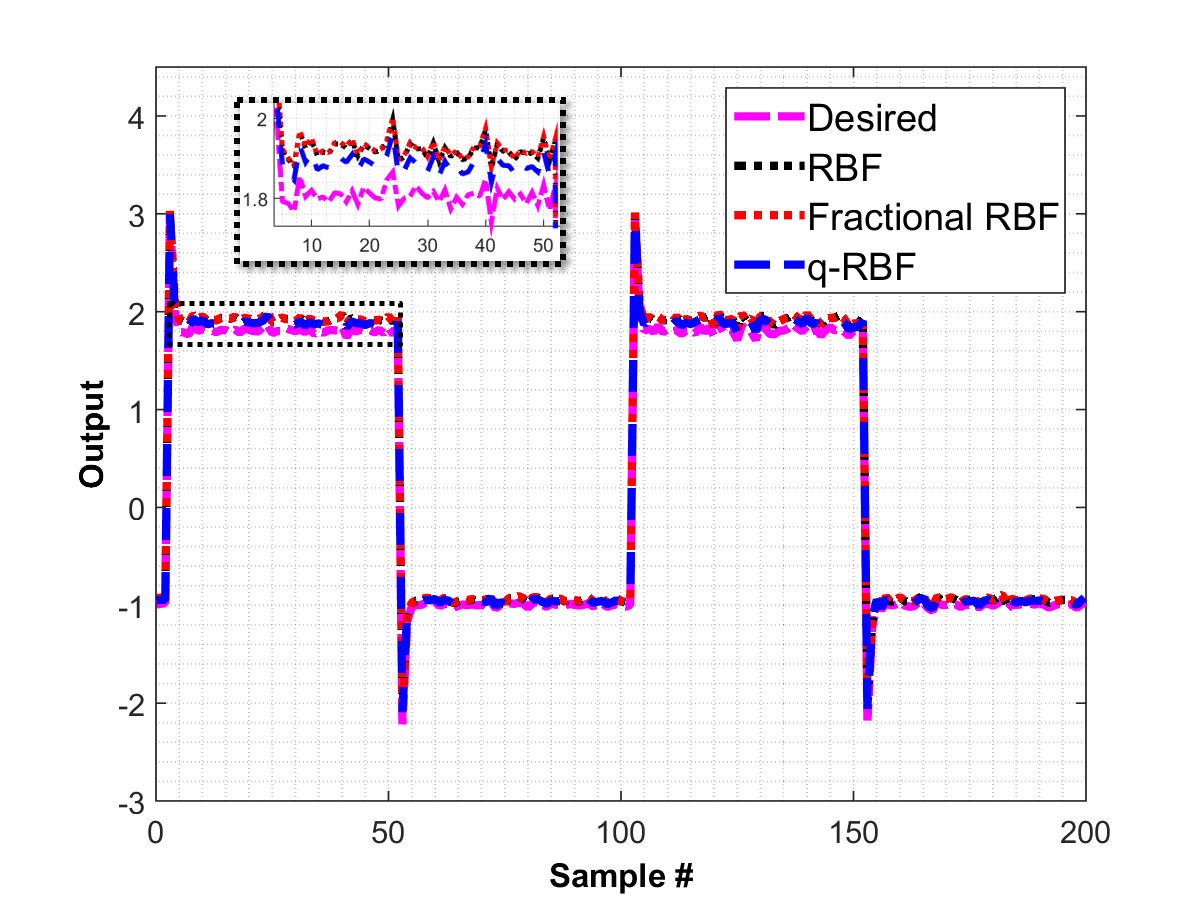}
	\end{center}
	\caption{Comparison of model and actual output for non-linear system identification problem}
	\label{fig:comparison}
\end{figure}

To better understand the origin of performance, we track the mean value of $q$ at each epoch. 
Note that the mean value of $q$-parameter reduces with the increase in the fitness of the RBFNN. This is well matching with the sensitivity analysis shown in section \ref{sec:sensitivity_analysis}. The comparison of the estimated output for the proposed algorithm with RBFNN and $f$-RBFNN during test  has been in shown Fig. \ref{fig:comparison}. 


\begin{figure}[!ht]
	\begin{center}
		\centering
		\includegraphics[width=8cm]
		{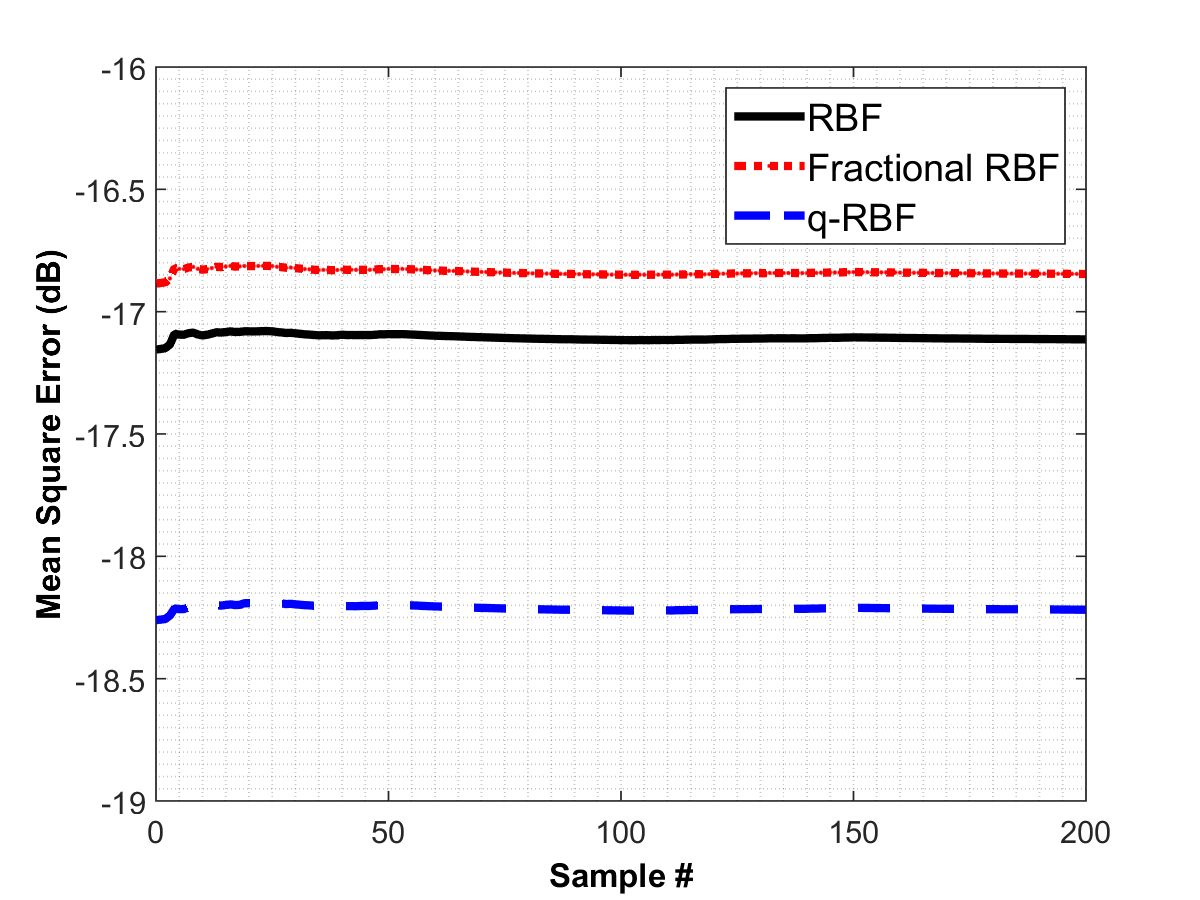}
	\end{center}
	\caption{MSE behavior in  during testing for non-linear system identification problem}
	\label{fig:test_mse}
\end{figure}

The MSE for the test has also been measured as shown in Fig. \ref{fig:test_mse}. Similar test conditions were retained for other architectures. It is observed that the proposed method achieves an MSE value of $18.20$ dB which better than those of RBF, FRBF, which achieved $16.85$ dB, $17.10$ dB respectively.

\subsection{Hammerstein model non-linearity estimation}

The proposed adaptive learning rate has been applied to estimate the static non-linearity present in the Hammerstein model. The simulation has been conducted using a Hammerstein model of a non-linear heat exchange coil along with linear dynamics as mentioned in \eqref{equ: hammerstein}. The proposed model is evaluated to estimate the model and the results are compared with the conventional RBFNN and $f$-RBFNN.
\begin{equation}
\label{equ: hammerstein}
\begin{split}
    &a(t)=-m_1r(t)+m_2r^2(t)-m_3r^3(t)+m_4r^4(t),\\
    &c(t)=n_1c(t-1)+n_2c(t-2)+n_3a(t)+h(t)
\end{split}
\end{equation}

\begin{figure}[!ht]
	\begin{center}
		\centering  
		\includegraphics[width=8cm]
		{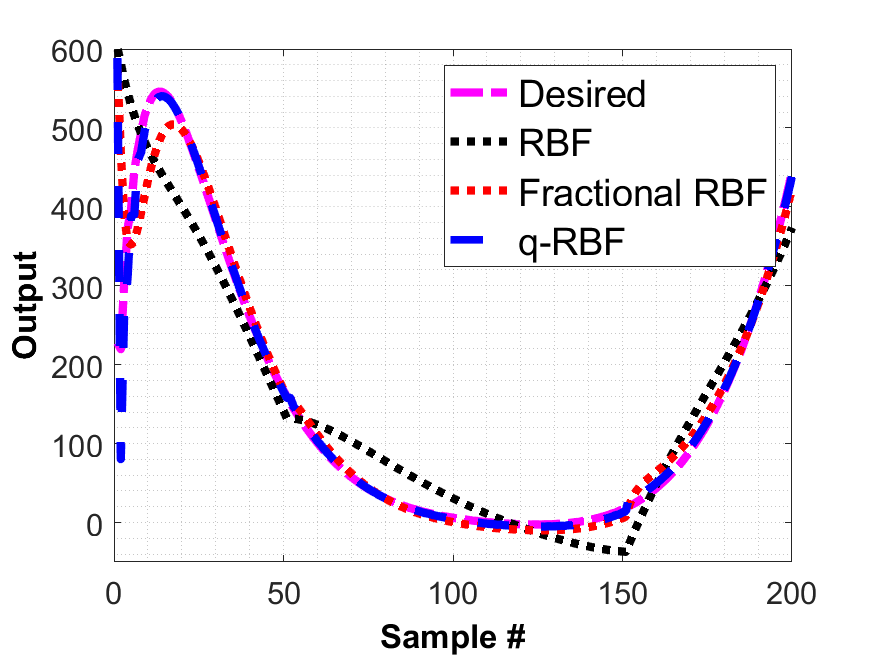}
	\end{center}
	\caption{Comparison of model and actual output for Hammerstein model non-linearity estimation}
	\label{fig:prob2_comparison}
\end{figure}

\begin{figure}[!ht]
	\begin{center}
		\centering
		\includegraphics[width=8cm]
		{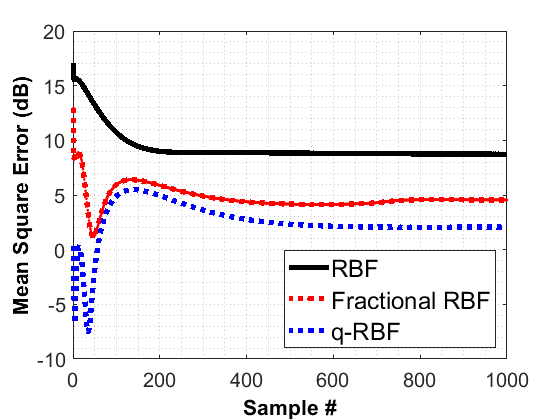}
	\end{center}
	\caption{MSE behavior during testing of Hammerstein model non-linearity estimation}
	\label{fig:prob2_MSE_test}
\end{figure}

The input $r(t)$ of the heat exchanger model is fed to the system to first get an intermediate result $a(t)$, which is further utilized for the output evaluation $c(t)$. The system is specified with some disturbance that is denoted by $h(t)$. For experimental setup the parameters of $m_1, m_2 \dots m_4$, $n_1, n_2, n_3$ and $h$ have to be selected for the output evaluation. The simulation using the proposed and conventional algorithms was carried out by keeping same values of all the parameters. We used the values of $31.549, 41.732, 24.201, 68.634$  for $m_1, m_2 \dots m_4$, $0.4,  0.35\  and\  0.15$ for $n_1, n_2, n_3$ and $0.1$ for the noise signal. The input signal is comprised of values ranging from $-2 to 2$ with a step size of $0.2$ constituted to generate $201$ samples of training.  
Fig. \ref{fig:prob2_comparison} depicts the plot of the desired and estimated results using the proposed and conventional algorithms where it can be observed that the proposed algorithm maps the desired output better than conventional ones. The tuning parameters of RBFNN, $f$-RBFNN, and proposed $q$-RBFNN are empirically opted in order to achieve the optimal results during training, and absolute value of gradient of the fractional update term is used to avoid complex values. There are several hyper-parameters in RBFNN and $f$-RBFNN algorithm including the mixing parameter $\alpha$ of gradient in $f$-RBFNN, which  is set to be $0.5$, learning rates $\eta$ and $\eta_\nu$ were both set to $1\times10^{-2}$, the fractional derivative power $\nu=0.9$ is used. \\
Efficacy of the proposed method is demonstrated using MSE curves in Fig. \ref{fig:prob2_MSE_test} which clearly shows the superior performance of the proposed method. In particular the MSE values achieved by RBFNN, $f$-RBFNN and proposed $q$-RBFNN are $8.85$dB, $4.56$dB and $2.15$dB respectively.


\subsection{Estimation of Non-linear MIMO System}
We consider a non-linear MIMO system with $2$ inputs and $2$ outputs to evaluate the efficacy of the proposed approach. The MIMO system in eq. \eqref{equ: mimo} has been utilized for estimation using the proposed method and the results are compared with the performances of conventional RBFNN and $f$-RBFNN.

\begin{multline}\label{equ: mimo}
    c_1(t)=m_1r_1(t)+m_2c_1(t-1)-m_3r_1(t-2)\\ +m_4cos(m_5r_2(t))+e^{-||r_1(t)||},\\
    c_2(t)=n_1r_2(t)-n_2r_2(t-1)+n_3r_2(t-2)\\+n_4sin(n_5r_1(t)) 
\end{multline}
The inputs and outputs of the system are represented by $r_1(t)$, $r_2(t)$ and $c_1(t)$, $c_2(t)$ respectively. 

\begin{figure*}[!ht]
	\begin{center}
		\centering  
		\includegraphics[width=18cm]
		{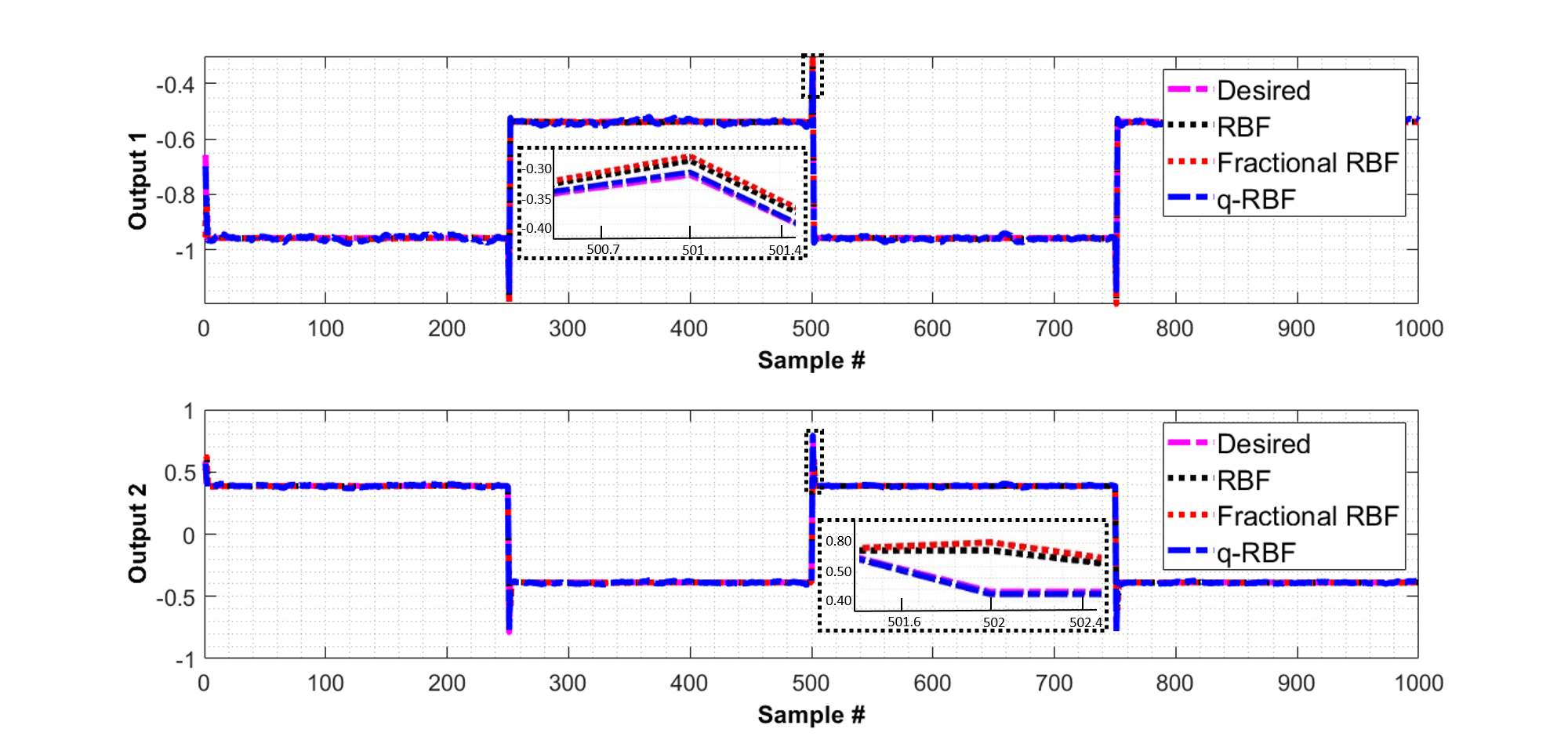}
	\end{center}
	\caption{Comparison of model and actual output for non-linear MIMO system}
	\label{fig:prob3_comparison}
\end{figure*}

The simulated results have been compared with the desired and estimated outputs of $q$-RBFNN, $f$-RBFNN and conventional RBFNN as shown in Fig. \ref{fig:prob3_comparison}. 

The values of parameters in \eqref{equ: mimo} are selected to be $0.21, -0.12, 0.3, -0.6\ and\ 0.5$  for $m_1, m_2 \dots m_5$ respectively, and  $0.25, -0.1, -0.2, 1.2\  and\ 0.2$ for $n_1, n_2, \dots n_5$ respectively while $\beta$, $\gamma$ and $q_{\max}$ of time-varying $q$-RBFNN are chosen to be $1$,$10$, and $10$ respectively.
The input the system is same as employed in \ref{secHNI}, except in case of MIMO where the inputs are $r_1(t)$ and $r_2(t)$, duplicated input streams were utilized for training of algorithm. 



There are several hyper-parameters in RBFNN and $f$-RBFNN algorithm including the mixing parameter $\alpha$ of gradient in $f$-RBFNN, which is set to be $0.5$, learning rates $\eta$ and $\eta_v$ are both set to $1\times10^{-3}$, the fractional derivative power $\nu=0.9$ is used. 
Fig. \ref{fig:prob3_comparison} depicts the desired and estimated results for the proposed and conventional algorithms where the proposed method is found to outperform the competitive approaches. The proposed method has achieved a lowest MSE of $-16.9$dB in $108$ iterations, compared to $-16.3$dB in $380$ iterations of RBFNN and $-16.4$dB in $370$ iterations, which signifies the superiority of the proposed method. The same has been depicted in Fig. \ref{fig:prob2_MSE_test}.



\begin{figure}[!ht]
	\begin{center}
		\centering
		\includegraphics[width=8cm]
		{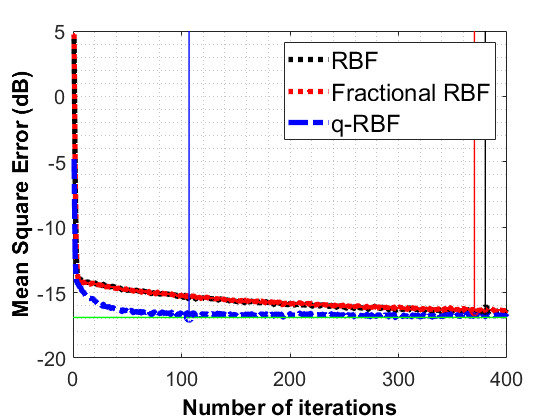}
	\end{center}
	\caption{MSE behavior during testing of non-linear MIMO system}
	\label{fig:prob3_MSE}
\end{figure}

\subsection{Chaotic Time Series Prediction}
Herein we perform the experiment of prediction of a chaotic time series which is a commonly used  quantitative model in signal processing. To determine the effectiveness of the proposed method, we consider a Mackey-Glass series which can be modeled as delayed differential equation \eqref{eq: cts}.
\begin{equation}
\label{eq: cts}
    \frac{dr(t)}{dt} = \frac{nr(t-\tau)}{1+r(t-\tau)^{10}} - mr(t)
\end{equation}

Here $r(t)$ is a time series of interval $t$=1,2,3,....,3000 derived from eq. \eqref{eq: cts} by performing sampling of the curve r(t) at the intervals of one second. $n$ and $m$ are the coefficients having values of 0.2 and 0.1 respectively, $\tau$ = 20 and $r(t-\tau)$ = 0 for $(\tau \geq t \geq 0)$. 

\begin{figure}[!ht]
	\begin{center}
		\centering  
		\includegraphics[width=8cm]
		{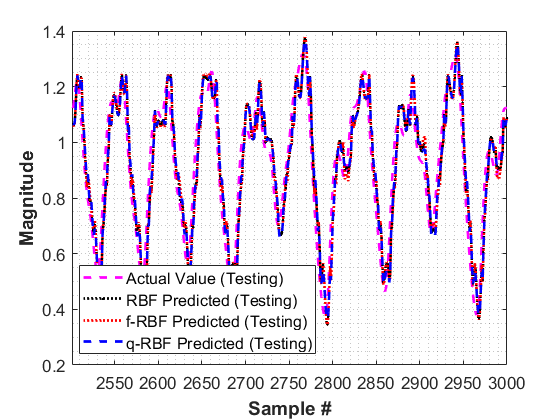}
	\end{center}
	\caption{Comparison of model and actual output for chaotic time series prediction}
	\label{fig:prob4_comparison_test}
\end{figure}

Furthermore, the training dataset achieved an SNR of $30dB$ upon introducing the white gaussian noise. In order to simulate for the proposed $q$-RBFNN, the samples from $100 \leq t \leq 2500$ has been considered to train the model. While, the samples lie in the range $2500 \leq t \leq 3000$ have been utilized for  testing purpose. For comparison with the proposed method, the conventional RBFNN and $f$-RBFNN are also trained using the similar parameters, the results of which are shown in Fig. \ref{fig:prob4_comparison_test}.

\begin{figure}[!ht]
	\begin{center}
		\centering
		\includegraphics[width=8cm]
		{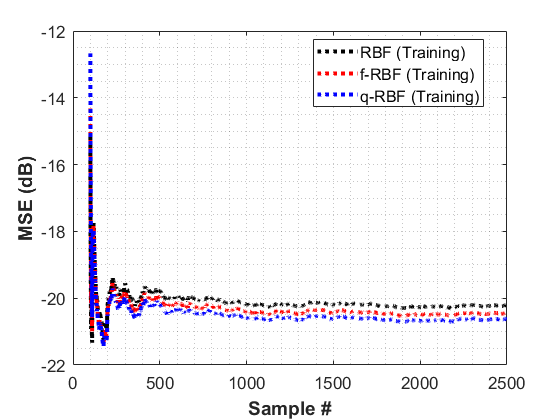}
	\end{center}
	\caption{MSE behavior during training of chaotic time series prediction model}
	\label{fig:prob4_MSE_train}
\end{figure}

\begin{figure}[!ht]
	\begin{center}
		\centering
		\includegraphics[width=8cm]
		{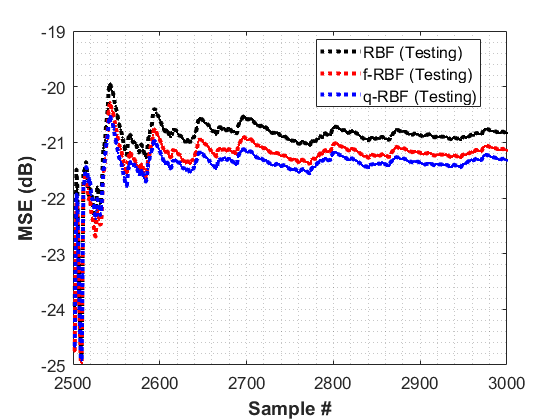}
	\end{center}
	\caption{MSE behavior during testing of chaotic time series prediction model}
	\label{fig:prob4_MSE_test}
\end{figure}

There are several hyper-parameters in RBFNN and $f$-RBFNN algorithm including the mixing parameter $\alpha$ of gradient in $f$-RBFNN, which is set to be $0.5$, learning rates $\eta$ and $\eta_v$ are both set to $1\times10^{-3}$, the fractional derivative power $\nu=0.5$ is used. K-means clustering has been employed to select the centers of the RBF and the algorithm is trained for $100$ epochs. The mean results reported in Fig. \ref{fig:prob4_MSE_train} depicts the MSE curves of training where as the mean results of MSE during achieved during test  are shown in Fig. \ref{fig:prob4_MSE_test}. From the results, it can be observed that the proposed $q$-RBFNN outperforms the conventional RBFNN, and $f$-RBFNN by achieving an average MSE of $-21.4$dB, while the conventional RBFNN and $f$-RBFNN achieve the average MSE of $-20.8$dB and $-21.1$dB respectively.


\section{Conclusion} \label{conclusion}
A quantum calculus based derivative  method is employed to propose a gradient descent based learning algorithm for RBFNN named as $q$-RBFNN. The proposed method exploits complementary properties of  quantum $q$ parameter which works on secant of the function taking larger steps to reach minima resulting in faster convergence. The proposed $q$-RBF is examined for the optimal solution in a system identification problem and an optimal Wiener solution is thus obtained. The transient behaviour and stability bounds of the learning rate (step-size) of the $q$-RBF is also derived and validated through computer simulations on a pattern classification problem. Using the analysis, results an adaptive framework for $q$-parameter is designed. The proposed time-varying $q$-RBFNN method is proven to be more promising and has shown to  outperform the conventional RBFNN on the problem of nonlinear system identification. The proposed method can be further improved by incorporating the evolutionary learning and sophisticated stochastic methods to train optimal models.

\section*{Acknowledgment}
Syed Saiq Hussain acknowledges the support of HEC, Pakistan under Indigenous Ph.D. Fellowship Program (PIN 417-29746-2EG4-120).

\bibliographystyle{IEEEtran}
\bibliography{Reference}

\end{document}